\documentclass[letterpaper]{article} 
\usepackage[]{aaai25}  
\usepackage{times}  
\usepackage{helvet}  
\usepackage{courier}  
\usepackage[hyphens]{url}  
\usepackage{graphicx} 
\usepackage{natbib}  
\usepackage{caption} 
\usepackage{algorithm}
\usepackage{algorithmic}

\usepackage{newfloat}
\usepackage{listings}
\DeclareCaptionStyle{ruled}{labelfont=normalfont,labelsep=colon,strut=off} 
\floatstyle{ruled}
\newfloat{listing}{tb}{lst}{}
\floatname{listing}{Listing}
\nocopyright 

\title{X-SQL: Expert Schema Linking and Understanding of Text-to-SQL with Multi-LLMs}

\author {
    Dazhi Peng
}
\affiliations{
    Carnegie Mellon University\\
    dazhip@alumni.cmu.edu
}

\usepackage{bibentry}

\begin{document}

\maketitle

\begin{abstract}
With Large Language Models' (LLMs) emergent abilities on code generation tasks, Text-to-SQL has become one of the most popular downstream applications. Despite the strong results of multiple recent LLM-based Text-to-SQL frameworks, the research community often overlooks the importance of database schema information for generating high-quality SQL queries. We find that such schema information plays a significant or even dominant role in the Text-to-SQL task. To tackle this challenge, we propose a novel database schema expert with two components. We first introduce X-Linking, an LLM Supervised Finetuning (SFT)-based method that achieves superior Schema Linking results compared to existing open-source Text-to-SQL methods. In addition, we innovatively propose an X-Admin component that focuses on Schema Understanding by bridging the gap between abstract schema information and the user's natural language question. Aside from better learning with schema information, we experiment with Multi-LLMs for different components within the system to further boost its performance. By incorporating these techniques into our end-to-end framework, X-SQL, we have achieved Execution Accuracies of 84.9\% on the Spider-Dev dataset and 82.5\% on the Spider-Test dataset. This outstanding performance establishes X-SQL as the leading Text-to-SQL framework based on open-source models.

\end{abstract}

%

\section{Introduction}\label{Introduction}
Text-to-SQL, the task of directly generating SQL code from natural language questions, has garnered significant attention with the advent of LLMs because of its potential to enable both professional and non-professional users to conduct efficient and accessible SQL data analysis \cite{qin2022survey}.

Despite the shared goal of generating executable code, Text-to-SQL is distinct from basic code generation tasks due to its dependence on database schema information. It is impossible to write an accurate SQL query without knowledge of the database schema. However, databases can range from containing a few tables to several hundred or even thousands, particularly in enterprise scenarios. A human data analyst may need to invest hours or even days to retrieve and comprehend the tables they are working with before they can begin writing the actual SQL code. This requirement presents significant challenges for LLM-based Text-to-SQL tasks. For instance, consider a database with candidate tables such as \texttt{Address}, \texttt{Staff}, \texttt{Lessons}, and \texttt{Customers}, along with a user's question:


\begin{quote}
\textit{``When did the staff member named Janessa Sawayn join the company?''}
\end{quote},
to generate the correct SQL query, it is essential to first determine the necessary tables. If we input the entire database schema into the LLMs, it will not only significantly raises the cost of inference tokens \cite{openai_pricing} but also increases the likelihood of hallucinations and errors in the generated SQL query \cite{liu2024lost}. The challenge of filtering relevant tables based on the user's question has been previously identified and known as Schema Linking in the literature \cite{pourreza2024din, li2024pet}.


Although existing LLM-based Text-to-SQL works \cite{pourreza2024din, talaei2024chess, dong2023c3, li2024pet}, involves some form of Schema Linking, the primary focus has been on in-context-learning-based approaches \cite{pourreza2024din, li2024pet, dong2023c3}. The only LLM-based Text-to-SQL study we know of that employs a training-based strategy is MAC-SQL \cite{wang2023mac}. However, MAC-SQL fine-tuned LLMs on a joint task that included Schema Linking, SQL generation, and Debugging. As a result, none of the existing studies have explored a dedicated LLM fine-tuning method for Schema Linking that could naturally enhance an LLM's ability to select the correct table more effectively.


To address this gap, we introduce \textbf{X-Linking}, a Schema Linking method based on Supervised Fine-Tuning (SFT). This approach specifically aims to improve LLMs' table selection capabilities by focusing training on the Schema Linking task. This dedicated training is crucial because LLMs typically do not develop the ability to retrieve relevant tables during their pre-training phase. Furthermore, during inference we employ a self-consistency strategy \cite{wang2022self}, a widely-used technique in LLM inference that involves shuffling inputs and aggregating outputs.

To demonstrate the effectiveness of X-Linking, we selected three state-of-the-art (SOTA) Text-to-SQL methods: DIN-SQL, PET-SQL, and MAC-SQL \cite{pourreza2024din, li2024pet, wang2023mac}. We evaluated the performance of their Schema Linking modules using only open-source models. Our findings show that X-Linking provides a significant advantage over existing Schema Linking methods, improving the overall accuracy of the end-to-end Text-to-SQL pipeline by 7\%.



Beyond simply retrieving the necessary tables, we also discovered that the accuracy of the final Text-to-SQL output is highly dependent on a deep understanding of the linked tables and their respective columns. In a typical data team, a data engineer often pauses to thoroughly comprehend the database schema, as this information is generally abstract. For instance, even if we successfully identify the required \texttt{Staff} table, it may still be challenging for LLMs to understand the necessary columns: \texttt{first\_name}, \texttt{last\_name}, and \texttt{date\_joined\_staff}, and to directly generate SQL queries based on them. We are the first to recognize and articulate this disconnect between SQL table definitions and user queries as the Schema Understanding problem. To address this issue, we introduce a novel component called \textbf{X-Admin}, which functions like a database administrator. X-Admin clarifies the extracted database schema in detailed natural language before feeding it into the LLMs for SQL query generation. Our results show that X-Admin can enhance the overall performance of the end-to-end Text-to-SQL system by a significant 1.7\%.


\begin{figure*}[t]
\centering
\includegraphics[width=1\linewidth]{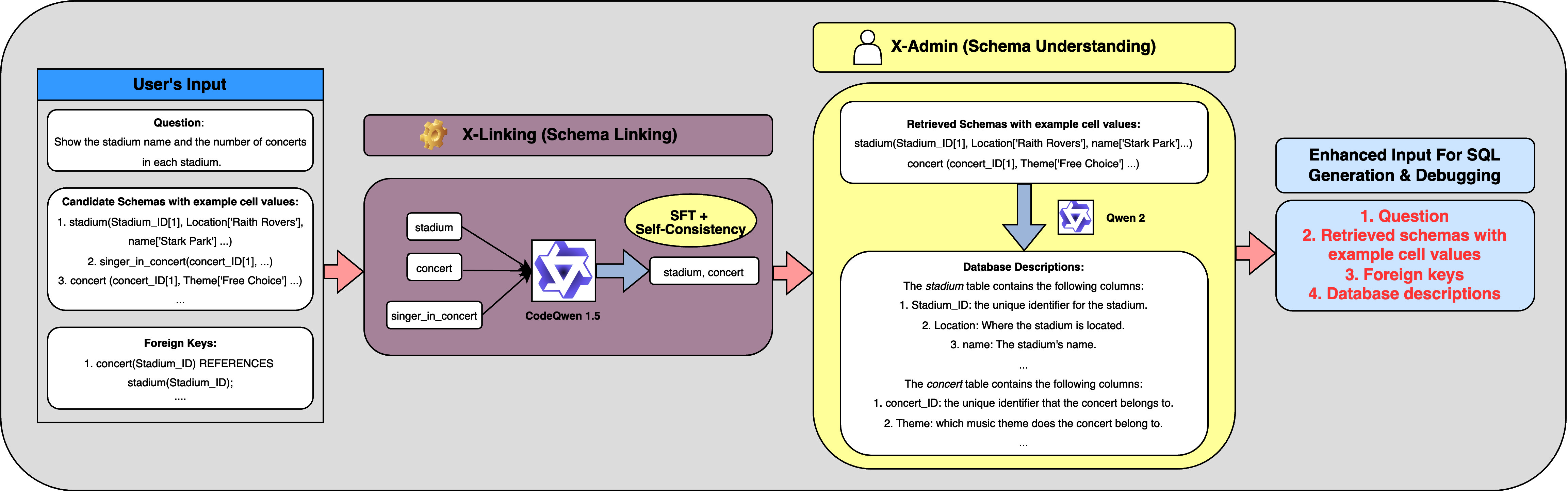} 
\caption{X-SQL's architecture. The candidate database schema is first filtered by X-linking. After that, X-Admin adds natural language descriptions to the linked table schema. Finally, we generate SQL queries with all this information and attempt to fix the queries if they execute with errors. The LLMs setup is based on the best Spider-Test result.}
\label{fig:X-sql}
\end{figure*}

In addition to the challenges posed by database schema, another unexplored area in the Text-to-SQL field is the use of Multi-LLMs. Given that Multi-LLMs have demonstrated significant potential in facilitating collaboration among various components within complex systems, as noted in recent studies \cite{feng2024don, shen2024small}, we believe that employing different LLMs for specific components within the Text-to-SQL system could be beneficial. By conducting experiments with four state-of-the-art coding and generic LLMs on components of the X-SQL system, we have established that a \textbf{Multi-LLMs-based} X-SQL system can achieve performance improvements. Specifically, a 1.3\% increase on the Spider-Dev dataset and a 2.2\% increase on the Spider-Test dataset—compared to a system that uses a single LLM.


To summarize, our contributions are as follows:
\begin{itemize}
\item We introduced X-Linking, a Schema Linking component based on SFT that outperforms the Schema Linking modules of three widely-used Text-to-SQL frameworks, achieving state-of-the-art results with a 7\% improvement.

\item We identified the challenge of Schema Understanding and have developed a novel component called X-Admin. This component enhances the X-SQL system by explaining the extracted database content in detailed natural language, effectively bridging the gap between abstract schema information and the user's queries. This innovation results in a significant performance boost of 1.7\% for the X-SQL system.

\item We have analyzed and confirmed the advantages of employing Multi-LLMs across various components within the X-SQL system, which led to a 2.2\% improvement. When utilizing only open-source models, the comprehensive X-SQL system outperforms other solutions, achieving the highest scores on the Spider-Dev (84.9\%) and Spider-Test (82.5\%) benchmarks.
\end{itemize}

\section{Related Work}
Recent years have witnessed significant advancements in the Text-to-SQL field, transitioning from early rule-based systems \cite{stratica2005using} to sophisticated Seq2Seq model-based machine-translation-like approaches \cite{zhong2017seq2sql}. The advent of LLMs has further propelled the development of Text-to-SQL frameworks. The past LLM-based Text-to-SQL works can be categorized into the following areas:

\subsection{In-context Learning Approaches}
DIN-SQL \cite{pourreza2024din} first breaks the Text-to-SQL task into four modules: Schema Linking, decomposition, generation, and self-correction, utilizing few-shot and well-designed prompts for each module. DIN-SQL has shown strong performance, particularly on challenging questions within the Spider \cite{yu2018spider} benchmark. However, C3 \cite{dong2023c3} identified that DIN-SQL's few-shot input context length often surpassed 10,000 tokens because of the many few-shot examples. To mitigate this, C3 introduced an effective zero-shot prompting method with clear templates and tips for model bias calibration, thereby enhancing performance.

In addition, DAIL-SQL \cite{gao2023text} evaluated the effectiveness of various prompt templates and introduced DAIL-Selection, an advanced dynamic few-shot example selection mechanism. DAIL-SQL was also the first work to assess open-source models' performance on Text-to-SQL tasks. MCS-SQL \cite{lee2024mcs} proposed a strategy similar to self-consistency \cite{wang2022self}, but by modifying candidate prompts' semantic structures and employing a Multiple-Choice selection mechanism. PET-SQL \cite{li2024pet} offered a comprehensive review of different prompt formats and proposed a Schema Linking method based on generating preliminary SQL queries and filtering relevant tables/columns from candidate SQL queries. PET-SQL also implemented a Multi-LLMs-based cross-consistency mechanism. All these works achieved notable success with closed-source models like GPT-4 \cite{achiam2023gpt} but were less effective with open-source models.

\subsection{Training-based Approaches}
RESDSQL \cite{li2023resdsql} trained a cross-encoder for the Schema Linking task and fine-tuned a Seq2Seq model by treating the SQL generation as a task similar to Machine Translation. CHESS \cite{talaei2024chess} achieved strong results on BIRD \cite{li2024can} by introducing an advanced cell value retrieval method and fine-tuning an SQL Generation LLM.

\subsection{Multi-Agents Approaches}
Recent research has also explored Multi-Agents systems, where LLMs are assigned specific roles and tools to work together to solve the Text-to-SQL challenge. MAC-SQL \cite{wang2023mac} introduced a Multi-Agents strategy with Selector, Decomposer, and Refiner agents and used either GPT-4 or their custom fine-tuned SQLLlama-7B models as agents' backbone LLMs. Inspired by MAC-SQL, SQLFixAgent \cite{cen2024sqlfixagent} focused on the refiner steps, generating multiple fixed query candidates using RubberDuck debugging.

\subsection{Other Approaches}
Beyond in-context learning and Multi-Agents methods, TA-SQL \cite{qu2024before} proposed generating Pandas API functions (replaced with symbolic representations) and then translating them to SQL code. This approach also demonstrated strong results using the GPT-4 model.

\section{X-SQL Framework}

The X-SQL framework comprises three components, as depicted in Figure \ref{fig:X-sql}. The first component, \textit{X-Linking}, identifies and extracts relevant database schema and foreign key information based on the user's question. Subsequently, the \textit{X-Admin} component translates the extracted abstract schema information into natural language, providing context that enhances the comprehensibility of the subsequent components. Finally, the \textit{SQL Generation \& Debugging} component formulates the SQL query using the insights provided by the X-Linking and X-Admin components and attempts to resolve any errors that arise after the query is executed against the actual database. On top of the three components, we also assess the performance of using different LLMs as the backbone model for X-SQL's different components and observe a significant improvement. The subsequent sections provide a detailed description of each component.

\subsection{X-Linking (Schema Linking)}
To address the accuracy limitations in existing Text-to-SQL frameworks and overcome the challenge that Schema Linking is not learned during LLM pre-training, we implement X-Linking, a Schema Linking module, through a dedicated SFT training process.

Specifically, we introduce a novel SFT method based on Q-LoRA \cite{dettmers2024qlora}. We carefully curate the dataset for SFT Schema Linking training and define the training objective as follows. The input comprises a prompt that concatenates a set of candidate table schemas \(S \) from a database, foreign key information \(K \), and a user query \(Q \). With a chosen LLM \(M \), the training objective for Schema Linking SFT is to maximize the likelihood that the LLM \(M \) correctly generates the appropriate table names \(T \). To simplify the training objective, we structure the target output  \(T \) as a sequence of table names, rather than the entire linked schema. Mathematically, the training process can be expressed as: \[
\max_{T} P(T \mid M(S, K, Q))
\] We adhere to the standard SFT practice by training on the cross-entropy loss between \(T \) and the generated table names \(T' \). Example test data is provided in Figure \ref{fig:input_example} and the complete Schema Linking prompt is detailed in Appendix A.

During inference, to mitigate the ordering bias of LLMs \cite{pezeshkpour2023large}, we randomly shuffle the order of candidate schemas \(S\) and foreign keys \(K\), generating multiple outputs. We then take the union of the output linked table names generated by the LLM \(M \) as the final result \(T' \). 

A detailed case study of X-Linking is presented in the Schema Linking component in Figure \ref{fig:X-sql}.

\begin{figure}[t]
\centering
\includegraphics[width=0.8\linewidth]{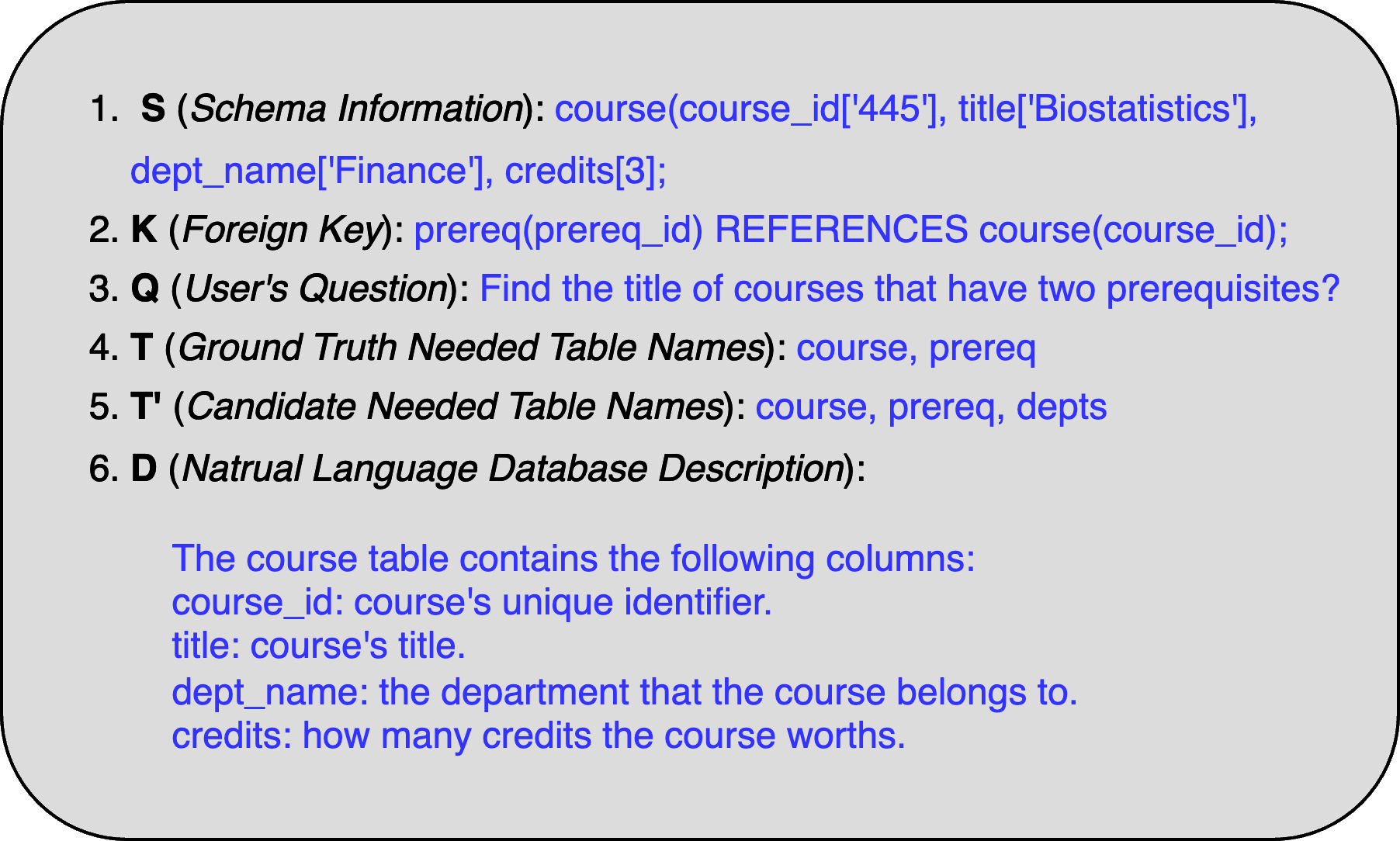}
\caption{Input Glossary Examples}
\label{fig:input_example}
\end{figure}
\subsection{X-Admin (Schema Understanding)}

The second crucial part of X-SQL's expert schema learning process is X-Admin, the Schmea Understanding component. After filtering out irrelevant database tables, a significant challenge remains: the abstract nature of database schema definitions. As illustrated in Figure \ref{fig:input_example}, the database schema \(S \) diverges markedly from the natural language or coding data typically encountered during LLM pre-training. This gap between abstract table definitions and user's natural language question has been overlooked by previous Text-to-SQL research \cite{pourreza2024din, gao2023text}.

\begin{figure}[t]
\centering
\includegraphics[width=0.8\linewidth]{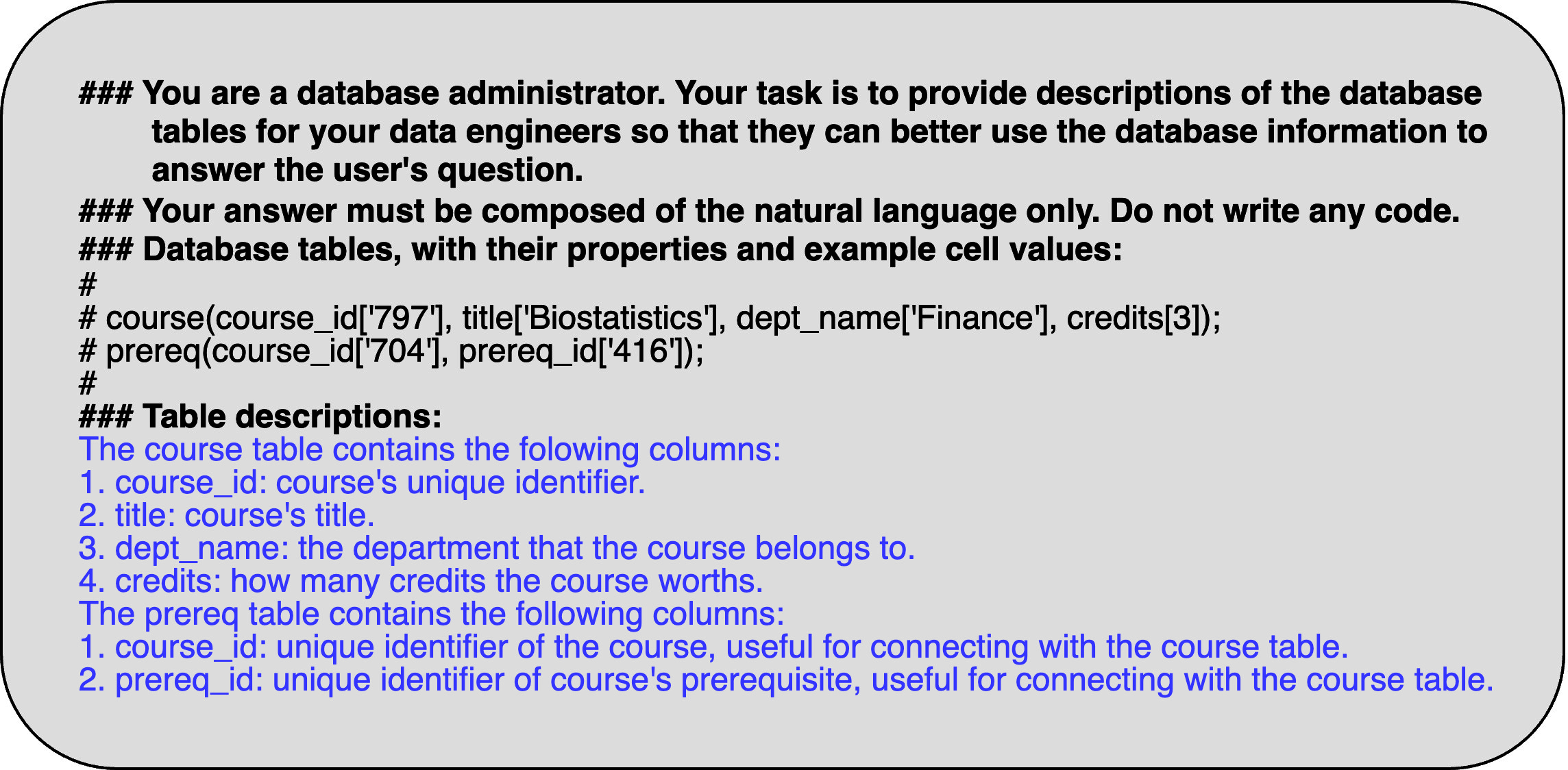} 
\caption{X-Admin (Schema Understanding) Prompt}
\label{fig:database_admin_prompt}
\end{figure}

To address this challenge, we introduce a novel Text-to-SQL component, X-Admin, which can also be seamlessly integrated into all Text-to-SQL frameworks. X-Admin assigns a specialized ``database expert" role to LLMs through carefully crafted role prompting. We opted not to pursue fine-tuning, as we believe that generating natural language explanations is a fundamental capability developed during LLM pre-training. Specifically, given linked schema from X-Linking, X-Admin uses natural language to elucidate the meaning of each table column based on their names and sample column values. Additionally, X-Admin provides hints to subsequent SQL Generation components on how certain columns might be used to connect different tables. For example, as shown in Figure \ref{fig:database_admin_prompt}, the abstract column \texttt{prereq(prereq\_id)} can be translated into a more comprehensible definition: \texttt{"unique identifier of course's prerequisite, useful for connecting with the course table"}. Such descriptions not only simplify the schema definition but also clarify the relationship between the \texttt{course} and \texttt{prereq} tables.
Our experiments demonstrate that X-Admin offers even greater benefits than the debugging component, which is considered essential in most Text-to-SQL frameworks. This superiority can be attributed to X-Admin's alignment with the role of an expert in real-world Software Engineering or Data Engineering teams. Such a human expert typically possesses an in-depth understanding of the codebase or database structure and can provide detailed explanations of each component and its interconnections. Therefore, X-Admin's expertise can enhance the contextual information available for subsequent stages of the X-SQL pipeline and bridge the gap between technical database structures and their practical applications. By integrating both X-Linking and X-Admin components, the X-SQL system achieves a level of database comprehension comparable to that of human experts, significantly mitigating the challenges that LLMs face when generating SQL queries from abstract database schemas.

\subsection{SQL Generation \& Debugging}
To make a complete system, X-SQL also has the SQL Generation \& Debugging components that adhere to standard practices established in previous Text-to-SQL research \cite{pourreza2024din, li2024pet, wang2023mac}, which we briefly outline here. Utilizing the linked schema information \(S^{\prime}\) from X-Linking, foreign keys \(K\), and generated database descriptions \(D\) from X-Admin, we input the user's query \(Q\) into the LLM to produce the corresponding \(SQL\) code. The SQL Generation prompt is provided in Appendix A.

Next, the generated \(SQL\) is executed on the SQLite database. If the execution proceeds without generating an error message \(E\), the original \(SQL\) is retained. If an error \(E\) occurs, the previously used information, along with \(E\), is reintroduced into the LLM \(M\) for a one-round debugging. The Debugging prompt is presented in Appendix A.

\subsection{Multi-LLMs-based Text-to-SQL System}
In addition to the proposed X-Linking and X-Admin components, our work is the first to explore the impact of applying Multi-LLMs to different components within the Text-to-SQL system. Drawing inspiration from the idea that a diverse workforce can enhance a company's productivity \cite{saxena2014workforce}, we hypothesize that the various components of the Text-to-SQL system function similarly to different roles within a company, while the diverse backbone LLMs represent candidates with varied educational and cultural backgrounds.

To thoroughly assess the impact of Multi-LLMs on the end-to-end Text-to-SQL's performance, we conduct extensive experiments using four SOTA open-source LLMs, including both general-purpose and coding-specific models, across all components of the X-SQL system. Our results demonstrate that the Multi-LLMs approach significantly enhances X-SQL's performance. Notably, we find that using different LLMs for the debugging component consistently improves X-SQL's performance. One possible explanation is that LLMs tend to favor content generated by themselves, as noted in \cite{panickssery2024llm}. We therefore hypothesize that leveraging different backbone LLMs for various Text-to-SQL components will not only help X-SQL, but also benefit other Text-to-SQL works.

\begin{table*}
    \centering
    \begin{tabular}{lllll}
        \hline
        \textbf{Frameworks} & \textbf{Spider-Dev} \( \mathbf{R_e} \) & \textbf{Spider-Dev \( \mathbf{R_s} \)} & \textbf{Spider-Test \( \mathbf{R_e} \)} & \textbf{Spider-Test \( \mathbf{R_s} \)} \\
        \hline
        DIN-SQL (CodeQwen1.5-7B-Chat) & 0.778 & 0.934 & 0.753 & 0.898 \\
        MAC-SQL (SQLLlama-7B) & 0.781& 0.949 & 0.779 & 0.926 \\
        PET-SQL (CodeLlama-34B + SQLCoder-34B) & 0.864 & 0.967 & 0.84 & 0.943 \\
        X-Linking (CodeQwen1.5-7B-Instruct) & \textbf{0.922} & \textbf{0.971} & \textbf{0.898} & \textbf{0.969}
    \end{tabular}
    \caption{Schema Linking Results}
\label{schema_linking_all}
\end{table*}

\section{Experiments}
The experiments are structured as follows: We first outline the datasets, metrics, and models used to evaluate the Schema Linking modules and the end-to-end Text-to-SQL systems. Subsequently, we compare the performance of X-Linking and X-SQL against other popular open-source methods. We then dive into X-Linking's backbone models comparison and X-Linking's effects compared to no and Oracle Schema Linking. Finally, we analyze the contributions of different components within X-SQL and the effects of applying Multi-LLMs to the system. All experiments are conducted three times, with the average results reported.

For all inference experiments, we design our prompt templates based on the SQL-Tailored prompt (a template involving optimization rules, example column values, and foreign keys) proposed in PET-SQL \cite{li2024pet}. We further refine the SQL-Tailored prompt by integrating its database properties (schema definition) section with its example column value section. Detailed descriptions of all prompt templates are provided in Appendix A.

\subsection{Datasets}
All experiments are conducted using the Spider dataset \cite{yu2018spider}, a large-scale, cross-domain, and widely-recognized Text-to-SQL benchmark. The Spider dataset consists of 8,659 entries in the training split and 1,034 entries in the development split, covering over 200 databases. Additionally, the recent release of the test split includes 2,147 entries across 34 databases \cite{li2024pet}.

\subsection{Metrics}
\subsubsection{Schema Linking}
To evaluate the effectiveness of retrieving relevant tables based on the user's question—specifically, the Schema Linking task—we employ the same \( R_e \) and \( R_s \) table recall metrics introduced in PET-SQL \cite{li2024pet}. Specifically, \( R_e \) indicates the percentage of questions where the linked table names \(T'\) exactly match the correct table names \(T\), while \( R_s \) reflects the percentage of questions where all \(T\) are included as a subset of \(T'\). In both cases, higher values indicate better performance.

\subsubsection{End-to-end Text-to-SQL}
To assess the correctness of the SQL queries generated by the end-to-end Text-to-SQL system, we focus on the widely used Execution Accuracy (\textit{Ex}) metric. \textit{Ex} assigns a value of 1 to a generated SQL query only when its execution result perfectly matches the ground truth. This metric is critical as it directly reflects the correctness of SQL query execution.

\begin{table}
    \centering
    \begin{tabular}{lll}
        \hline
        \textbf{Frameworks} & \textbf{Spider-Dev EX} & \textbf{Spider-Test EX} \\
        \hline
        \multicolumn{3}{l}{\footnotesize \textit{Trained with SQL Code Generation on Spider-Train}} \\
        RESDSQL & 84.1 & 79.9 \\
        MAC-SQL & 76.3 & 70.6 \\
        \hline
        DIN-SQL& 75.6 & - \\
        PET-SQL & 82.2 & - \\
        X-SQL & \textbf{84.9} & \textbf{82.5} \\
    \end{tabular}
    \caption{End-to-end Text-to-SQL results}
\label{end2end_all}
\end{table}


\subsection{Pre-trained Models Compared}
To ensure a fair comparison, we exclusively experiment with instruction-tuned language models in this paper:
\begin{itemize}
    \item \textit{Qwen2-7B-Instruct} \cite{yang2024qwen2}
    \item \textit{llama-3-sqlcoder-8B} \cite{llama3sqlcoder8b}
    \item \textit{deepseek-coder-7b-instruct-v1.5} \cite{guo2024deepseek}
    \item \textit{CodeQwen1.5-7B-Chat} \cite{codeqwen}
\end{itemize}
For the X-Linking component's SFT model, we also explore the potential of the Encoder-Decoder model \textit{Flan-T5-XXL} (11B) \cite{chung2024scaling} to assess the feasibility of Seq2Seq models in schema linking.

\begin{table}
    \centering
    \begin{tabular}{lll}
        \hline
        \textbf{Models} & \textbf{\( \mathbf{R_e} \)} & \textbf{\( \mathbf{R_s} \)} \\
        \hline
        Flan-T5-XXL (11B) & 0.722 & 0.746\\
        \hline
        llama-3-sqlcoder-8B & 0.713 & \textbf{0.999} \\
        deepseek-coder-7b-instruct-v1.5 & 0.843 & 0.882\\
        Qwen2-7B-Instruct & 0.874 & 0.928\\
        CodeQwen1.5-7B-Instruct & \textbf{0.891} & 0.948\\
        \hline
    \end{tabular}
    \caption{Schema Linking SFT models on Spider-Dev}
\label{schema_linking_models}
\end{table}

\begin{table*}
    \centering
    \begin{tabular}{llll|l}
        \hline
        \textbf{Models} & \textbf{Train Context Length} & \textbf{w/o SL} & \textbf{w/ X-Linking} & \textbf{w/ Oracle Schemas}\\
        \hline
        llama-3-sqlcoder-8b & 8K & 61.2 & 62.6 (+1.4) & 66.2 \\
        Qwen2-7B-Instruct & \textbf{131K} & 71 & 75.3 (+4.3) & 75.7 \\ 
        deepseek-coder-7b-instruct-v1.5 & 4K & \textbf{75.1} & 76.7 (+1.6) & 78.5 \\
        CodeQwen1.5-7B-Chat & 64K & 74.3 & \textbf{81.6 (+7.3)} & \textbf{84.4}  \\
        \hline
    \end{tabular}
    \caption{X-Linking's effects on Spider-Dev SQL Generation}
\label{sl_sql}
\end{table*}

\subsection{Schema Linking Results}
We compare X-Linking with the Schema Linking methods of three prominent Text-to-SQL frameworks on the Spider-Dev and Spider-Test datasets. Specifically, we include the following prior works:

\begin{itemize}
    \item \textbf{DIN-SQL} \cite{pourreza2024din}: This method employs a custom prompt with many-shot examples to guide LLMs in retrieving relevant tables. We reproduce DIN-SQL's Schema Linking using CodeQwen1.5-7B-Instruct, as the original work utilized GPT-4.
    \item \textbf{PET-SQL} \cite{li2024pet}: This method introduces an alternative Schema Linking approach by asking LLMs to generate a preliminary SQL and filter schemas based on that SQL. We use PET-SQL's provided best result based on open-source models.
    \item \textbf{MAC-SQL} \cite{wang2023mac}: This method is the only work that utilizes a training-based Schema Linking method. However, MAC-SQL integrated different Text-to-SQL subtasks, such as Schema Linking and Debugging, into a shared training objective.
\end{itemize}

It is important to note that, although we have experimented with various LLMs for our Schema Linking model, we report only the best-performing result (achieved with CodeQwen1.5-7B-Chat) for simplicity in this section. In the ``X-Linking Ablation Studies'' section, we detail the selection process of the LLMs used for the Schema Linking SFT model.

As shown in Table \ref{schema_linking_all}, X-Linking consistently outperforms all existing Schema Linking methods across both \(R_e\) and \(R_s\) metrics on the Spider-Dev and Spider-Test datasets. Specifically, X-Linking demonstrates a 5.5\% boost in the \(R_e\) metric and a 2\% improvement in the \(R_s\) metric compared to the current best result. X-Linking also significantly surpasses the training-based MAC-SQL approach, underscoring the effectiveness of our proposed Schema Linking dedicated training method. With X-Linking, we not only greatly simplify the schema input for later Text-to-SQL components but also recall more relevant tables. These enhancements highlight the potential for further advancements in Schema Linking modules within current Text-to-SQL frameworks.

\subsection{End-to-end Text-to-SQL Results}
In addition to X-Linking, we evaluate the performance of our end-to-end X-SQL method against four leading Text-to-SQL approaches, using only open-source models as the backbone LLMs on the Spider-Dev and Spider-Test datasets:

\begin{itemize}
    \item \textbf{MAC-SQL} \cite{wang2023mac}: This method employs a Multi-Agent system comprising Selector, Decomposer, and Refiner agents. We use the results with SQLlama-7B, built on CodeLlama-7B \cite{roziere2023code} and fine-tuned across multiple Text-to-SQL subtasks.
    \item \textbf{DIN-SQL} \cite{pourreza2024din}: This method introduces the Decomposer module, which breaks down complex questions into multiple steps. We reproduce the results using CodeQwen1.5-7B-Chat, as the original DIN-SQL was evaluated with GPT-4.
    \item \textbf{PET-SQL} \cite{li2024pet}: This method leverages custom Text-to-SQL prompts and multiple LLMs for cross-consistency. We utilize the best-provided results based on a five-LLM ensemble.
    \item \textbf{RESDSQL-3B + NatSQL} \cite{li2023resdsql}: This training-based approach combines cross-encoder and Seq2Seq models with the intermediate representation NatSQL \cite{gan2021natural}. We use the its best result for analysis.
\end{itemize}

\begin{table}
    \centering
    \begin{tabular}{lc}
        \hline
        \textbf{Components} & \textbf{Spider-Dev Ex} \\
        \hline
        w/o Debugging & \(\downarrow\) 1.6 \\ 
        w/o X-Admin & \(\downarrow\) 1.7 \\
        w/o X-Linking & \(\downarrow\) 7.3 \\
        \hline
    \end{tabular}
    \caption{X-SQL Components Ablation Studies}
\label{components_ablation}
\end{table}

Following the simplicity principle, we present the best LLM setup for X-SQL. Details on how we configure X-SQL's Multi-LLMs method can be found in the section ``Multi-LLMs Ablation Studies''.

As shown in Table \ref{end2end_all}, X-SQL achieves SOTA end-to-end performance, surpassing all existing approaches. Notably, X-SQL outperforms RESDSQL and MAC-SQL (both explicitly trained for Schema Linking and SQL Generation) by 0.8\% on the Spider-Dev set and 2.6\% on the Spider-Test set, respectively. These results underscore the efficacy of our proposed X-Linking (a dedicated LLM SFT Schema Linking method) and X-Admin (schema understanding) components, which significantly enhance the model's contextual comprehension and its ability to generate more accurate SQL queries. Furthermore, X-SQL exceeds the performance of PET-SQL, the leading prompting-based approach, by 2.7\%. Interestingly, PET-SQL employs five distinct LLMs for cross-consistency inference within a single component, whereas X-SQL applies Multi-LLMs by assigning each component with a different backbone LLM. This approach underscores the significance of assigning specialized LLMs to different components within the Text-to-SQL system.




\begin{table*} 
    \centering
    \begin{tabular}{lll}
        \hline
        \textbf{X-Admin \& Debugging LLMs} & \textbf{Spider-Dev Ex} & \textbf{Spider-Test Ex}\\
        \hline
        CodeQwen1.5-7B-Chat + CodeQwen1.5-7B-Chat & 83.6 & 80.3 \\
        CodeQwen1.5-7B-Chat + deepseek-coder-7b-instruct-v1.5 & \textbf{84.9} & 81.7 \\ 
        Qwen2-7B-Instruct + llama-3-sqlcoder-8B & 83.4 & \textbf{82.5} \\ 
        \hline
    \end{tabular}
    \caption{Multi-LLMs Ablation Studies. Remember we have fixed CodeQwen-1.5-7B-Chat as the backbone LLMs for X-Linking and SQL Generation.}
\label{multi_llms_ablation}
\end{table*}

\subsection{X-Linking Ablation Studies}
\label{x_linking_ablation}
Given the significant advancements demonstrated by our X-Linking component, we conduct comprehensive experiments to deepen the understanding of its design choices and implementation.

\subsubsection{Backbone SFT Model Benchmarking}
To select the optimal backbone SFT model, we benchmark five SOTA instruction-tuned language models on the Schema Linking SFT task, with results presented in Table \ref{schema_linking_models}. Detailed information on the SFT and Q-LoRA hyperparameters can be found in Appendix B.
\begin{itemize}
    \item \textit{Flan-T5-XXL}: Despite its large parameter count as a Seq2Seq model, its performance in the Schema Linking SFT task is underwhelming (0.722 \(R_e\) \& 0.746 \(R_s\)).
    \item \textit{llama-3-sqlcoder-8B}: Although this model achieves a seemingly perfect \(R_s\) (0.999), the result is misleading due to its very low \(R_e\) (0.713). The \(R_s\) score will default to 1 if the entire database schema is input without any schema linking.
    \item \textit{deepseek-coder-7b-instruct-v1.5} \& \textit{Qwen2-7B-Instruct}: The performance of these models is inferior (3\% less) to that of \textit{CodeQwen1.5-7B-Instruct}.
\end{itemize}
Therefore, aiming to balance between \(R_e\) and \(R_s\), we select CodeQwen1.5-7B-Chat as the base model for our Schema Linking SFT task (0.891 \(R_e\) \& 0.948 \(R_s\)).

\subsubsection{X-Linking vs. w/o SL vs. Oracle Schemas}
To compare the impact of X-Linking, schema linking without any enhancements (inputting all tables), and Oracle Schemas (ground truth tables) on the accuracy of generated SQL queries, we applied these three linking scenarios to four different LLMs as a plug-and-play step before SQL query generation.

As shown in Table \ref{sl_sql}, X-Linking significantly enhances SQL generation performance across all four models compared to the scenario without Schema Linking scenarios. The performance of X-Linking also closely aligns with the results achieved using Oracle Schemas. Notably, the Qwen Family models exhibit greater benefits from X-Linking than the llama3-sqlcoder-8B and deepseek-coder-7b-instruct-v1.5 models, despite having longer training context lengths. Given the trend toward increasingly extended training contexts in LLMs, we anticipate that robust Schema Linking methods like X-Linking will become even more impactful for future LLMs.

\subsection{X-SQL Ablation Studies}
\subsubsection{Components Ablation Studies}
We conduct a detailed analysis to evaluate the contribution of each component within the X-SQL framework to the overall performance of the end-to-end system. To streamline the experiments within the Multi-LLM-based X-SQL system, we establish CodeQwen1.5-7B-Chat as the backbone LLM for X-Linking and SQL generation in all experiments, due to its superior performance as demonstrated in Tables \ref{schema_linking_models} and \ref{sl_sql}. Given the vast number of potential LLM combinations for each component in the X-SQL system, we present the results of the optimal LLM combination in Table \ref{components_ablation}. Comprehensive results, including setups for all LLM combinations across different components, are provided in Appendix C.

\begin{itemize}
    \item \textit{w/o Debugging}: First, we evaluate X-SQL's performance without the Debugging component, which is a standard component in all state-of-the-art Text-to-SQL systems. We observe a 1.6\% drop in the end-to-end system's performance.
    \item \textit{w/o X-Admin}: Next, we assess the system's performance without the X-Admin component. Interestingly, this leads to a 1.7\% drop, which is larger than the drop observed without the Debugging component. This finding highlights the importance of our proposed X-Admin Schema Understanding method, demonstrating that it is at least as crucial as the well-known Debugging approach.
    \item \textit{w/o X-Linking}: Finally, we test the system without the X-Linking component, which results in a significant 7.3\% drop in performance. This suggests that X-Linking is the most critical component in the X-SQL system.
\end{itemize}

\subsubsection{Multi-LLMs Ablation Studies}
To underscore the advantages of our proposed Multi-LLM-based X-SQL system over a single-LLM-based system, we compare the performance of our end-to-end system under two scenarios: (1) both the X-Admin and Debugging components employ the same CodeQwen1.5-7B-Chat LLM as the other components, and (2) one or both of the X-Admin and Debugging components utilize different LLMs. We focus on different LLM setups for the X-Admin and Debugging components because we have previously established CodeQwen1.5-7B-Chat as the backbone LLM for X-Linking and SQL generation. Due to space constraints, we present only the top-performing results for the Multi-LLM system in Table \ref{multi_llms_ablation}, with the complete experimental results detailed in Appendix C.

As shown in Table \ref{multi_llms_ablation}, the Multi-LLM-based X-SQL system consistently outperforms its single-LLM-based counterpart across both the Spider-Dev and Spider-Test datasets. Notably, utilizing different LLMs for the Debugging component, rather than relying solely on CodeQwen1.5-7B-Chat for both SQL generation and debugging, consistently yields superior performance. This finding supports our earlier hypothesis that LLMs may exhibit bias toward their own outputs and that incorporating different LLMs in the Debugging component effectively mitigates this issue.

\section{Conclusion}
In this paper, we present X-SQL, an Expert Schema Learning Text-to-SQL framework that leverages Multi-LLMs. Our novel Expert Schema Learning system, comprising the X-Linking and X-Admin components, enhances X-SQL's performance by 7.3\% and 1.7\%, respectively. Notably, X-Linking achieves state-of-the-art (SOTA) results compared to existing Text-to-SQL Schema Linking modules. Through the integration of a Multi-LLM-based system, X-SQL sets a new benchmark in the open-source-models-based Text-to-SQL field, achieving scores of 84.9 on the Spider-Dev dataset and 82.5 on the Spider-Test dataset. We hope this research will inspire further exploration and development in Text-to-SQL systems, particularly in the areas of stronger Schema Learning and Multi-LLM-based systems.

\bibliography{aaai25}

\section{Reproducibility Checklist}
\subsubsection{This paper}
\begin{itemize}
\item Includes a conceptual outline and/or pseudocode description of AI methods introduced: \textbf{Yes}
\item Clearly delineates statements that are opinions, hypothesis, and speculation from objective facts and results: \textbf{Yes}
\item Provides well marked pedagogical references for less-familiare readers to gain background necessary to replicate the paper: \textbf{Yes}
\end{itemize}

\subsubsection{Theoretical Contributions}
\begin{itemize}
\item Does this paper make theoretical contributions? \textbf{No}
\end{itemize}

\subsubsection{Datasets}
\begin{itemize}
\item Does this paper rely on one or more datasets? \textbf{Yes}
\item A motivation is given for why the experiments are conducted on the selected datasets \textbf{Yes}
\item All novel datasets introduced in this paper are included in a data appendix. \textbf{N/A}
\item All novel datasets introduced in this paper will be made publicly available upon publication of the paper with a license that allows free usage for research purposes. \textbf{N/A}
\item All datasets drawn from the existing literature (potentially including authors’ own previously published work) are accompanied by appropriate citations. \textbf{Yes}
\item All datasets drawn from the existing literature (potentially including authors’ own previously published work) are publicly available. \textbf{Yes}
\item All datasets that are not publicly available are described in detail, with explanation why publicly available alternatives are not scientifically satisficing. \textbf{N/A}
\end{itemize}

\subsubsection{Experiments}
\begin{itemize}
    \item Does this paper include computational experiments? \textbf{Yes}
    \item Any code required for pre-processing data is included in the appendix. \textbf{No}
    \item All source code required for conducting and analyzing the experiments is included in a code appendix. \textbf{No}
    \item All source code required for conducting and analyzing the experiments will be made publicly available upon publication of the paper with a license that allows free usage for research purposes. \textbf{Yes}
    \item All source code implementing new methods have comments detailing the implementation, with references to the paper where each step comes from \textbf{Yes}
    \item If an algorithm depends on randomness, then the method used for setting seeds is described in a way sufficient to allow replication of results. \textbf{No}
    \item This paper specifies the computing infrastructure used for running experiments (hardware and software), including GPU/CPU models; amount of memory; operating system; names and versions of relevant software libraries and frameworks. \textbf{Partial}
    \item This paper formally describes evaluation metrics used and explains the motivation for choosing these metrics. \textbf{Yes}
    \item This paper states the number of algorithm runs used to compute each reported result. \textbf{Yes}
    \item Analysis of experiments goes beyond single-dimensional summaries of performance (e.g., average; median) to include measures of variation, confidence, or other distributional information. \textbf{No}
    \item The significance of any improvement or decrease in performance is judged using appropriate statistical tests (e.g., Wilcoxon signed-rank). \textbf{No}
    \item This paper lists all final (hyper-)parameters used for each model/algorithm in the paper’s experiments. \textbf{Yes}
    \item This paper states the number and range of values tried per (hyper-) parameter during development of the paper, along with the criterion used for selecting the final parameter setting. \textbf{No}
\end{itemize}

\clearpage
\appendix
\section{Appendix}
\subsection{A. Prompts Templates}
\subsubsection{X-Linking}
\begin{figure}[h!]
\centering
\includegraphics[width=0.9\linewidth]{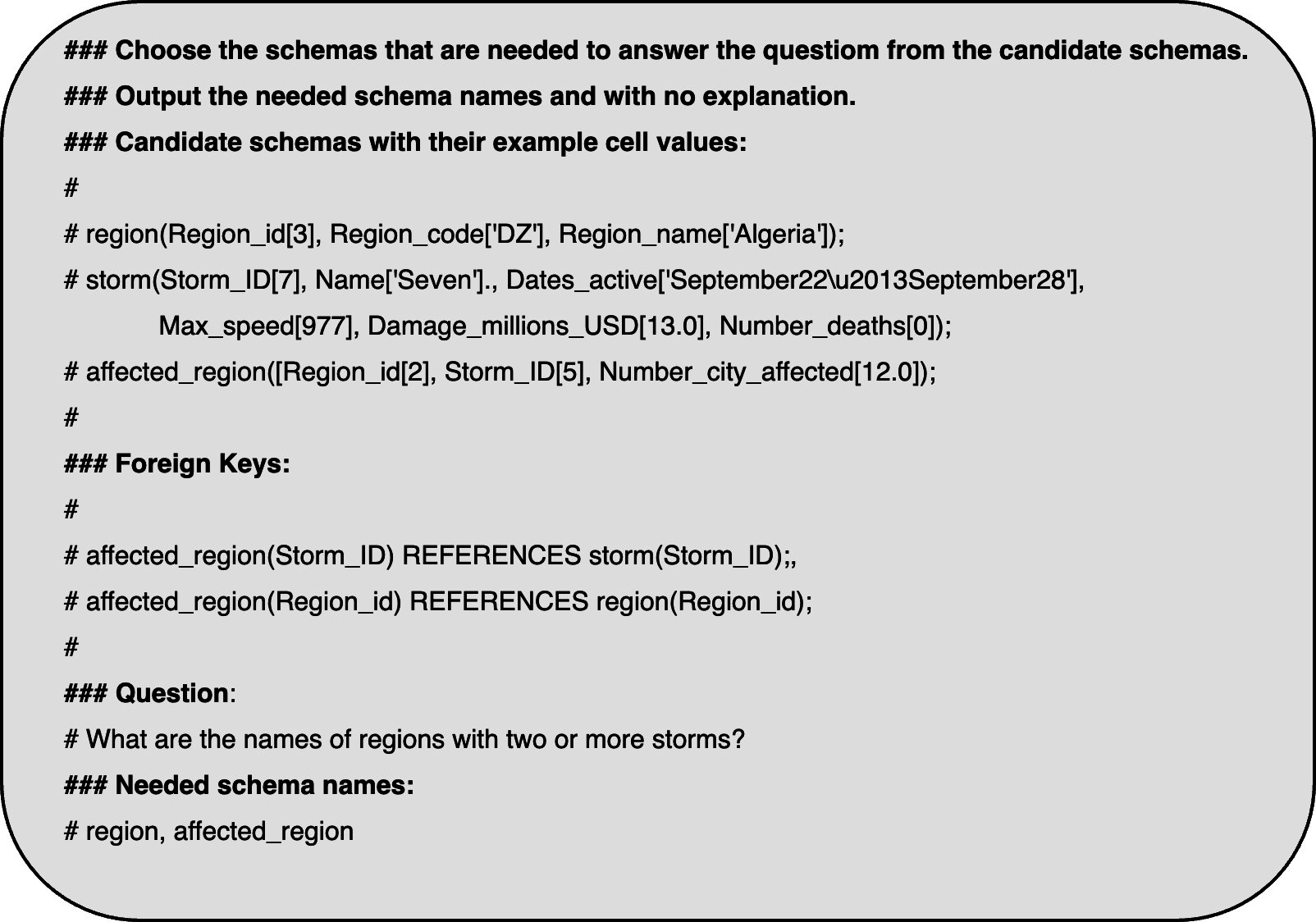} 
\caption{X-Linking Prompt Template}
\label{fig:schema_linking_prompt}
\end{figure}

As depicted in Figure \ref{fig:schema_linking_prompt}, the input for both SFT training and inference consists of the candidate schemas \(S\), foreign keys \(K\), and the user's query \(Q\). The specific implementations of SFT training and inference are detailed as follows:
\begin{itemize}
    \item \textbf{SFT Training}: Given the input, the training objective is to predict the ground truth for the linked table names \(T\), that are associated with the context following the instructions ``\#\#\# \textit{ Needed schema names}'' in Figure \ref{fig:schema_linking_prompt}.

   
    \item \textbf{Inference}: To incorporate our enhanced inference strategy, we randomly shuffle the order of the input candidate schemas \(S\) and foreign keys \(K\) five times, generating five distinct sets of output linked table names \(T'\). The final output, corresponding to the context following the instruction ``\#\#\# \textit{ Needed schema names}'' in Figure \ref{fig:schema_linking_prompt}, is derived by taking the union of these five sets.
    
\end{itemize}

\subsubsection{SQL-Generation}
\begin{figure}[h!]
\centering
\includegraphics[width=0.9\linewidth]{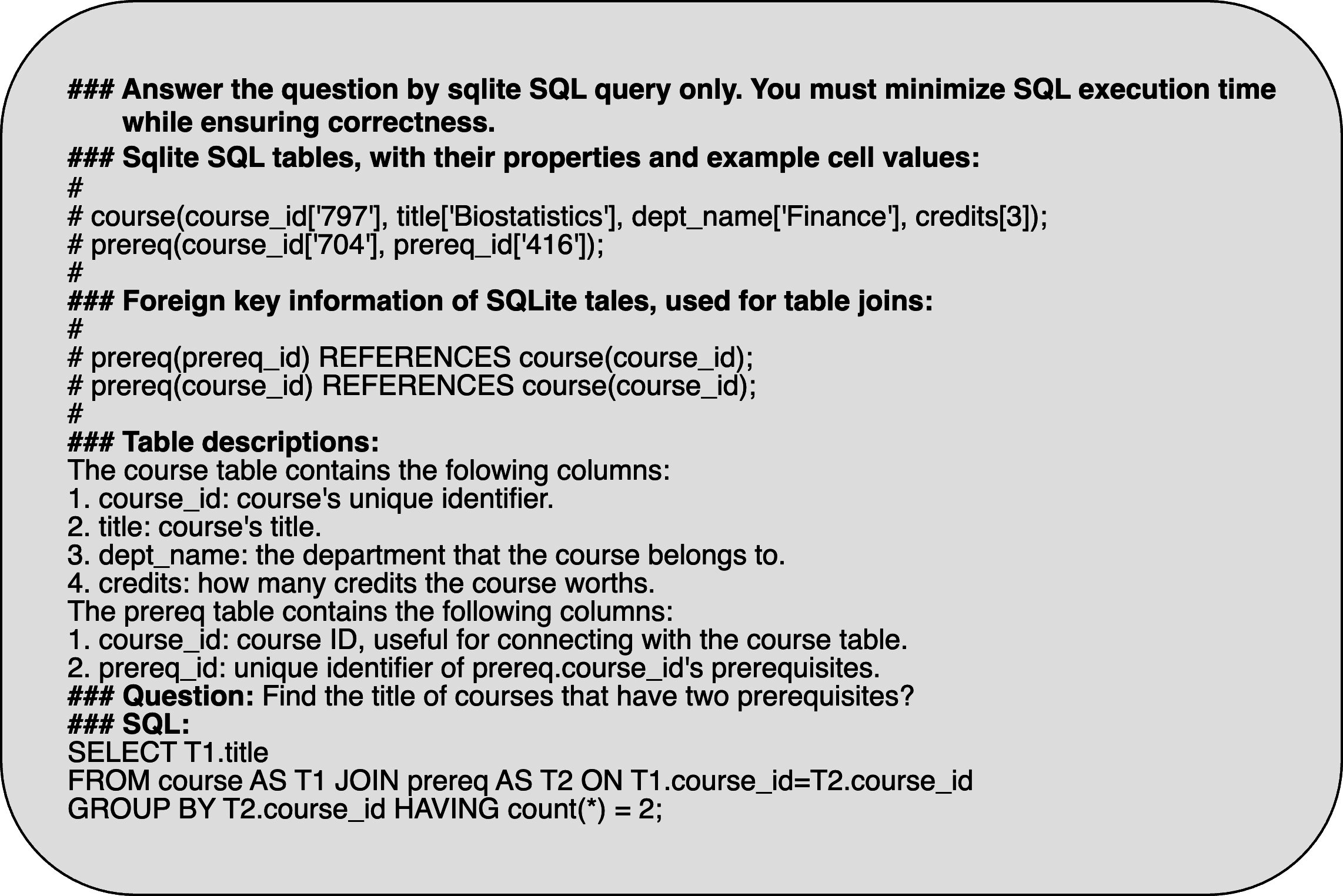} 
\caption{SQL Generation Prompt Template}
\label{fig:sql_generation_prompt}
\end{figure}

As depicted in Figure \ref{fig:sql_generation_prompt}, the SQL Generation prompt leverages the linked schema \(S'\) produced by X-Linking, the table descriptions \(D\) provided by X-Admin, the filtered foreign keys \(K'\) based on \(S'\), and the user's question \(Q\) to generate the candidate \(SQL\).

\subsection{Debugging}
\begin{figure}[h!]
\centering
\includegraphics[width=0.9\linewidth]{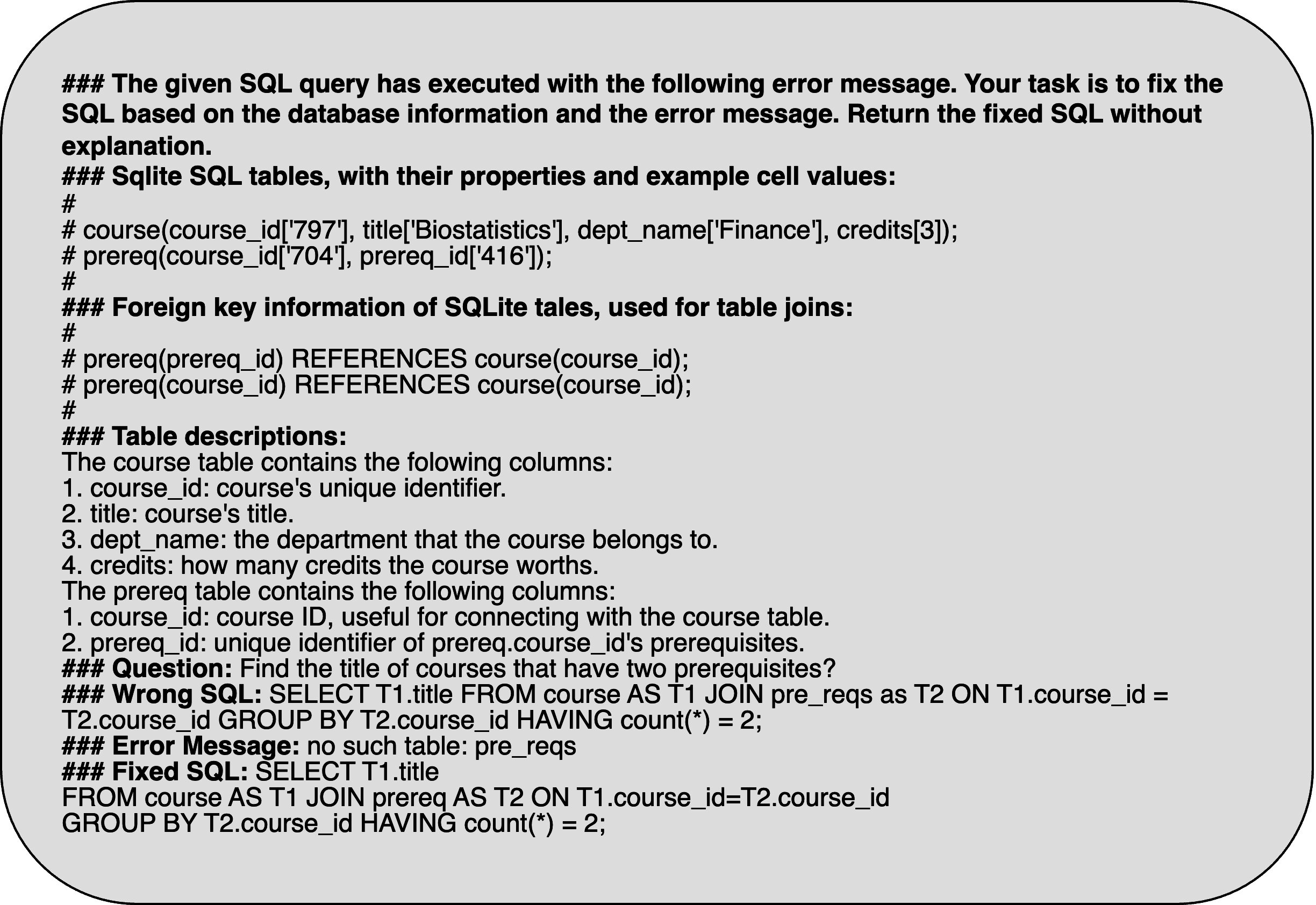} 
\caption{Debugging Prompt Template}
\label{fig:debugger_prompt}
\end{figure}

If the candidate \(SQL\) produces an error \(E\) during execution, we initiate the debugging process. As illustrated in Figure \ref{fig:debugger_prompt}, the candidate \(SQL\), along with linked schema \(S'\), foreign keys \(K\), error message \(E\), and database descriptions \(D\), are re-input into the LLM \(M\) for debugging.

\subsection{B. X-Linking Training Details}
We fine-tune our Schema Linking SFT model on 2 NVIDIA Tesla V100 GPUs, utilizing the HuggingFace Transformers, Accelerate, and Bitsandbytes packages.

\subsubsection{Common Hyperparameter Settings:}
This section details the training configurations employed for both the decoder-only models and the encoder-decoder model.
\begin{itemize}
    \item Training Precision: \texttt{fp16}
    \item LoRA Parameters: \texttt{Rank=128; Alpha=256, Target Modules=Q, K, V, Dropout=0.1}
    \item Learning Rate Scheduler: \texttt{constant}
    \item Train Batch Size: \texttt{32}
    \item Max Grad Norm: \texttt{0.3}
     \item Warm-up Ratio: \texttt{0.03}
\end{itemize}

\subsubsection{Decoder-Only Hyperparameter Settings:}
This section outlines the configurations applied to the decoder-only language models.
\begin{itemize}
    \item Learning Rate: \texttt{1e-4}
    \item Optimizer: AdamW \cite{loshchilov2017decoupled}
    \item Epoch: \texttt{1}
\end{itemize}

\subsubsection{Encoder-Decoder Hyperparameter Settings:}
Following the methodology described in the original Flan-T5 paper \cite{chung2024scaling}, we adapt our learning rate scheduler to use AdaFactor \cite{shazeer2018adafactor}. Furthermore, we increase the number of training epochs to mitigate the initially sub-optimal training loss.
\begin{itemize}
    \item Learning Rate: \texttt{5e-4}
    \item Optimizer: AdaFactor 
    \item Epoch: \texttt{3}
\end{itemize}

\subsection{C. Multi-LLMs Experiments}
This section presents the experimental results of integrating various LLMs within the end-to-end X-SQL system. Building on the earlier ``Experiments'' section, where CodeQwen-1.5-7B-Chat is fixed as the backbone LLM for the X-Linking and SQL Generation components, we now shift our focus to evaluating various LLMs for the X-Admin and Debugging components. For clarity, the following abbreviations are used for the LLMs tested:
\begin{itemize}
    \item \textit{CQ}: CodeQwen1.5-7B-Instruct
    \item \textit{LS}: llama-3-sqlcoder-8B
    \item \textit{DC}: deepseek-coder-7b-instruct-v1.5
    \item \textit{QW}: Qwen2-7B-Instruct
\end{itemize}
As demonstrated in Table \ref{multi_llms_ablation}, the optimal results on both the Spider-Dev and Spider-Test datasets are achieved by utilizing different LLMs for either or both the X-Admin and Debugging components, rather than exclusively relying on CodeQwen-1.5-7B-Chat. These findings underscore the effectiveness of our proposed Multi-LLMs strategy.

We present the best results, specifically the \texttt{CQ + DC} and \texttt{QW + LS} results, in the ``Multi-LLMs Ablation Studies'' section of the main paper.

\begin{table}[h]
    \centering
    \begin{tabular}{lll}
        \hline
        \textbf{X-Admin + Debugging} & \textbf{Spider-Dev} & \textbf{Spider-Test}\\
        \hline
        CQ + LS & 83.5 & 80.3 \\
        CQ + DC & \textbf{84.9} & 81.7 \\
        CQ + CQ & 83.6 & 80.3 \\
        CQ + QW & 83.4 & 80.3 \\
        \hline
        DC + LS & 84 & 81.1 \\
        DC + DC & 83.3 & 81.8 \\
        DC + CQ & 83.7 & 80.1 \\
        DC + QW & 83.4 & 79.7 \\
        \hline
        QW + LS & 83.4 & \textbf{82.5} \\
        QW + DC & 83.8 & 81.9 \\
        QW + CQ & 82.7 & 79.8 \\
        QW + QW & 83.5 & 80.6 \\
        \hline
        LS + LS & 82.4 & 81.7 \\
        LS + DC & 83 & 82 \\
        LS + CQ & 81.9 & 78.9 \\
        LS + QW & 80.9 & 79.3 \\
        \hline
    \end{tabular}
    \caption{Execution accuracies of all possible backbone LLM combinations for the X-Admin and Debugging components.}
    \label{table:multi_llms_all}
\end{table}


\subsection{D. X-Linking SFT Contributions}
\begin{table}[htbp!]
    \centering
    \begin{tabular}{lll}
        \hline
        \textbf{Strategies} & \textbf{\( \mathbf{R_e} \)} & \textbf{\( \mathbf{R_s} \)}  \\
        \hline
        3-shot ICL & 0.326 & 0.334\\
        SFT & 0.891 (\textbf{+0.565}) & 0.948 (\textbf{+0.614}) \\
        SFT + EI & 0.922 (+0.031) & 0.971 (+0.023) \\
        \hline
    \end{tabular}
    \caption{Schema Linking strategies on Spider-Dev}
\label{schema_linking_union}
\end{table}

This section highlights the importance of our proposed SFT Schema Linking strategy within the X-Linking component. As shown in Table \ref{schema_linking_union}, we compare Schema Linking performance across three scenarios: 3-shot in-context learning (3-shot ICL), SFT alone, and SFT combined with enhanced inference (SFT + EI). Notably, the SFT approach significantly improves Schema Linking effectiveness, yielding gains of 0.565 on \(R_e\) and 0.614 on \(R_s\). The enhanced inference strategy further contributes a modest increase of 0.031 on \(R_e\) and 0.023 on \(R_s\) increase. These results underscore the value of dedicated training for Schema Linking in our methodology.
\end{document}